\documentclass[11pt,a4paper]{article}
\usepackage[hyperref]{acl2020}
\usepackage{times}
\usepackage{latexsym}

\usepackage{microtype}

\usepackage{graphicx}
\usepackage{appendix}

\title{ESAN: Efficient Sentiment Analysis Network of A-Shares Research Reports for Stock Price Prediction}

\author{Tuo Sun, Wanrong Zheng, Shufan Yu, Mengxun Li, Jiarui Ou}
\date{Latest Revision: \today}

\begin{document}
\maketitle

\section{Introduction}

A shares, also known as domestic shares, are shares that are denominated in Chinese currency Renminbi and traded in the Shanghai and Shenzhen stock exchanges, as well as the National Equities Exchange and Quotations. Financial analysts, as a dominant occupation of the Chinese financial services industry, write millions of research reports to pore over data to identify opportunities or evaluate outcomes for business decisions or investment recommendations. According to statistics, in the last decade, around 550,000 analyst research reports have been written for A shares analysis. These reports will violently change the stock trend.

However, reading reports brings a lot of problems to stock price prediction work. On one side, since there are usually thousands of characters for each report, reading all the reports becomes a slow and laborious task for financial analysts.

On the other hand, the signals from these reports have a time effect. Factors are also known as features in Machine Learning tasks. In finance, a factor refers in particular to signals composed in time-series which is usually composed of two dimensions, stock dimension and time-series dimension. The factors generated by these signals will decay in 1 month. After that, it can hardly affect the market. Thus, a sentiment analysis of A-Shares research reports is extremely useful for financial analysts to save them from repetitive and low-efficiency labor.

\begin{figure}[t] 
    \centering
    \includegraphics[width=\columnwidth]{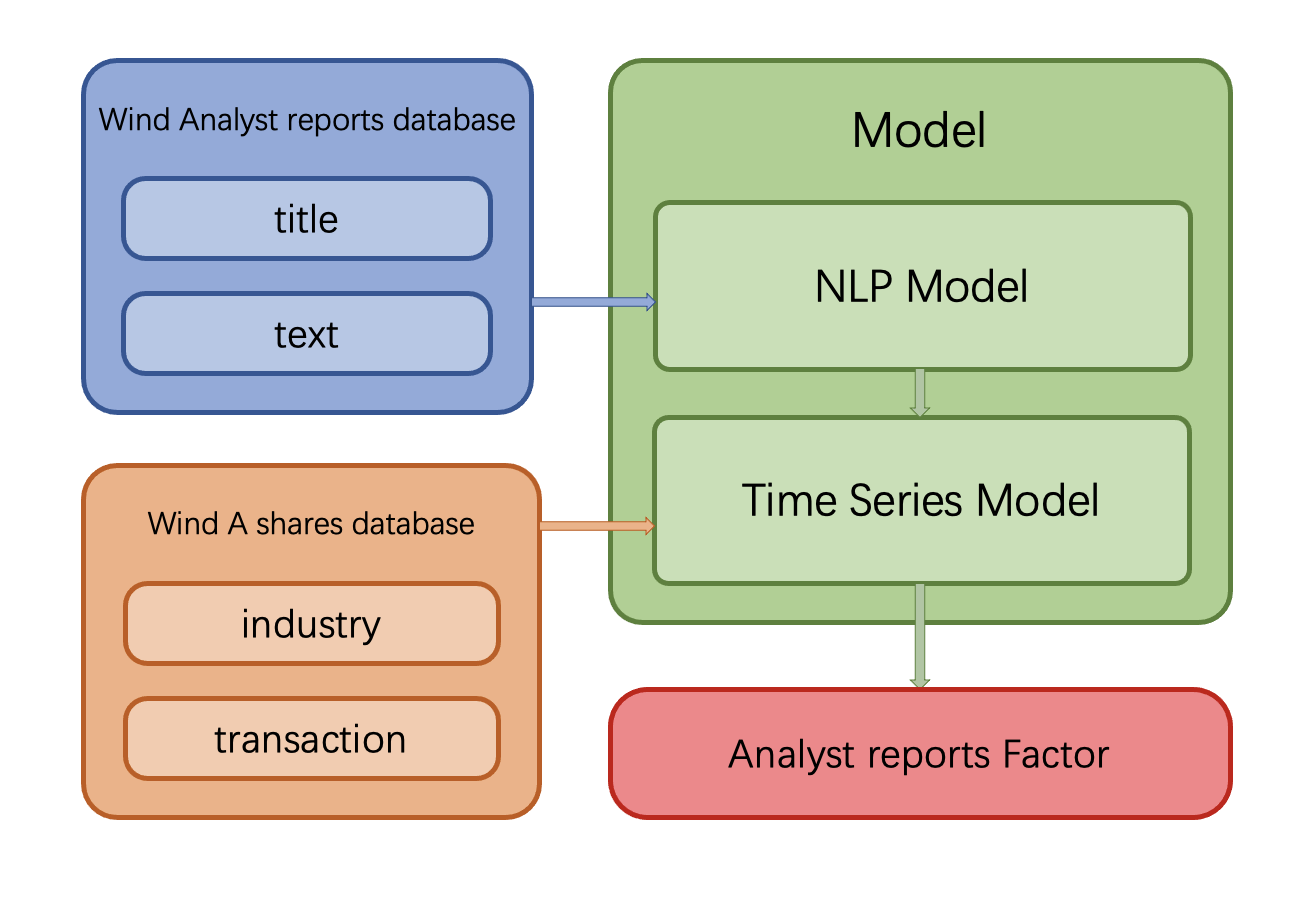}
    \caption{Overall Workflow of ESAN. The 1st module takes the report title and text snippets as input, and its outputs will be sent into 2rd module which is a time series model. The time series model combines industry and transaction information with sentiment analysis outputs and generates final analyst reports factor.}
    \label{fig:overall}
\end{figure}

In this paper, we are going to develop a natural language processing model to help us to predict stocks in the long term. The whole network includes two modules as shown in figure~\ref{fig:overall}. The first module is a natural language processing model which seeks out reliable factors from input reports. While the other is a time-series\cite{elliot2017time} forecasting model which takes the factors as input and aims to predict stocks earnings yield. To indicate the efficiency of our model to combine the sentiment analysis module and the time-series forecasting module, we name our method \textbf{ESAN}.

To sum up, the contributions of this article are three-hold:

(1) We propose the NLP + stock price prediction pipeline to address the time-cost problem in human financial analysis. Our final solution aims to generate a fast end-to-end sentiment analysis model and achieve superior performance to previous research.

(2) NLP module provides extraordinary ability to generate the stock dimension factors. We use RoBERTa\cite{liu2019roberta} as the NLP module, which is a robustly optimized BERT pretraining approach.

(3) The mean combinatory strategy of ESAN reaches 20\% annualized rate of return and 7.19\% RankIC. Compared to previous statistical analysis approaches, our pipeline is stable when handling time-series data and can prolong the time-efficiency of output factors.

\section{Methods}
\subsection{Related Work}
A lot of previous research has been done to predict the stock price utilizing NLP techniques. For example, Priyank Sonkiya’s team did experiments to predict the stock price of Apple Inc\cite{sonkiya2021stock}. Stock price prediction using BERT\cite{DBLP:journals/corr/abs-1810-04805} and GAN\cite{goodfellow2014generative}. They first performed sentiment analysis on news about Apple Inc through fine-tuned Bert. Afterwards, the sentiment score was combined with other features such as stock indexes and sent into a GAN or LSTM\cite{sak2014long} model to give the prediction. Additionally, Jaydip Sen performed a similar experiment to predict the stock price of NIFTY 50 by doing sentiment analysis on twitter and related news events\cite{2021}. However, both of these research only focus on predicting the stock price of a single company and therefore lack generalizability. Furthermore, the texts they used for sentiment analysis are from news and social networks rather than professional analysis, which could be an obstacle to reaching better performance.
In another previous research by Xiaoming Lin, he proposed a basic approach to predict the stock price of A shares\cite{huataizhengquan}. However, his model utilizes a binary classifier. With the manual tagging labels, he trains a model to predict the title of these analyst reports to generate a confidence for each of them. This model can get around 98\% accuracy. 
The network is trained for 5 epochs in total, with an initial learning rate of \(1\times 10 ^ {-5}\). This model can get around 0.9833 accuracy and 0.9762 AUC for the testset. However, such a model is not appropriate to be applied as factors. The meaning of the model is separated from financial markets. It only focuses on sentiment analysis of words instead of the return of investments.

\subsection{The Proposed Approach}
Following the descriptions in Xiaoming’s paper, we build and improve a model to analyze the title and abstracts of analyst research reports and predict return rates for stocks in A shares. Taking the reports as input, the NLP model generates the returns of each stock that we can use in a complex stock prediction model. Since the returns are continuous outputs, this means that our NLP model is essentially a regression model instead of a binary classification in Xiaoming’s original model. Also, Xiaoming’s model fine-tunes Bert to perform sentiment analysis for the topic of analysis reports and uses these intermediate results to make predictions on the stock prices. But to improve performance, we used the training time of the model by fine-tuning RoBERTa-tiny-clue instead of bert-base-chinese. To generate the returns as continuous labels, we select the next trade day of the release time of the last paper in each group. In order to research the timelines of the factors in this topic, we generate labels for various lengths of time regarding the trading days as we will discuss in the later Experiment section. Besides, we also generate the returns of the whole market in the same time period then we can calculate the difference of features between each stock and the whole market to get the exact return as the final feature. In this way, the final feature is more accurate without the influence of other stocks. Then, we can take the outputs from the above NLP model as input to the complex stock prediction model, which would give us a prediction of stocks earnings.

Finally, we then can provide an investment strategy such as equal rights investment (i.e., split all your investment equally to each stock) using the above outputs and calculate the profit from this strategy to see how well our model performs in real A shares stock market.

\subsection{Network Architecture}
To conclude, our whole pipeline consists of two main modules: firstly, a module of the natural language processing model which outputs the returns of each stock as continuous labels from the input reports as discussed in section 2.2, plus a second module of the complex stock prediction model that takes the returns as input and predicts stocks earnings yield. We then evaluate our model by applying an investment strategy based on the top selected stocks that have the highest predicted earnings from our pipeline to see how much actual profit we can gain using this strategy based on our own model.

\section{Experiments}
\subsection{Datasets}
\noindent \textbf{Wind Analyst Reports Database.}
We select 369581 of stock market analyst reports in A shares. We also record the release time of each date. However, in this data, we do not need a manual binary label for each article like the previous study. We will use numerical values from the second dataset as labels, thus we change from a classification problem to a regression problem.


\noindent \textbf{Wind A shares Database.}
The time-series database for the stocks. Our main goal is to calculate time periods of returns of each stock using this database. These time periods are set to 5, 10, 20, 40, 60 trade days.

We then combine two datasets mentioned above. However, the number of the reports for each stock is imbalanced, which means the model will concentrate on some special words for the popular and trending stocks. In order to solve this imbalanced problem, we grouped the reports in the adjacent release time for each stock. 
Besides, we calculate the active return of each stock, which is the absolute return of each stock minus the whole A share market’s average return.

Based on previous introduction and discussion, the stock data of the project mainly consists of time-series data. Therefore, we have four time split points with the format: yyyy-mm-dd, which are 2018-12-31, 2019-06-30, 2019-12-31,2020-06-30. And there are four rolling models, the training set is 80\% of the data before each time split point, the rest 20\% is the validation set and the data after each time split point is the test set. 

\noindent \textbf{Evaluation Protocols.}
We employ the RankIC for evaluation. In stock prediction, to get an annualized return rate is the most direct way to evaluate the model. However, there are a lot of strategies and policies to allocate assets to invest in a stock market. So in quantification, we usually use other metrics to evaluate the performance of the model. Therefore, we usually use RankIC as an evaluation method. 
In mathematics, it can also be called Spearman correlation coefficient. This kind of correlation coefficient is commonly used as a criterion of factors in Quantitative Analysis, though it has no gradient as a rank-like loss function. The RankIC tells us whether the ranks of our alpha values are correlated with the ranks of the future returns. If the future performance of the assets matched the expectations that was suggested by the alpha factor, then the information coefficient would be higher. Otherwise, it would be lower and possibly negative.

\noindent \textbf{Implementation Details.}
\emph{Model Backbones:} The architecture of the ESAN is demonstrated in figure \ref{fig:overall}. We have 4 layers of Transformer which consists of 4 attention heads and the hidden state is 312. The overall number of parameters is 7.5 million.
\emph{Training Hyper-parameters:} AdamW\cite{loshchilov2019decoupled} with an epsilon of \(1\times 10 ^ {-8}\) is adopted to update parameters. The network is trained for 10 epochs in total, with an initial learning rate of 0 and warms up to \(5\times 10 ^ {-5}\) at the first epoch. Learning rate is reduced from \(5\times 10 ^ {-5}\) to 0 gradually after one epoch. Mean squared error is used as loss function. We set the max sequence length to 500. 

\section{Results and Discussions}
In a previous research, Xiaoming Lin gave a basic approach to predict the binary label of A shares stock prices changes (Increasing or decreasing) based on the titles of analyst reports. His model reached 98\% accuracy. A corresponding stock selecting strategy would lead to 12.35\% annualized rate of return and 3.03\% RankIC. Even though 3.03\% RankIC is relatively small, the strategy still proves to be fruitful and could bring stable revenue. 

In order to investigate the timeliness of our method, we trained and tested our models several times using the analyst reports from different periods of time. The results show our model can not perform as well for predicting far-future stock prices as predicting near-future stock prices. When using a single model to make the investment decision, the annualized return rate drops dramatically after the 19th month past the time of training data. The most possible explanation of this phenomenon is the existence of some hot words that may contain strong signals for a given period but lose efficacy soon in the future. 

\begin{table}[hbt]
\centering
\resizebox{\columnwidth}{!}{%
\begin{tabular}{llll}
\hline
\multicolumn{1}{c}{Models} & \multicolumn{1}{c}{Training data start time} & \multicolumn{1}{c}{Training data end time} & \multicolumn{1}{c}{Test set RankIC}\\ \hline
1 & 20090101 & 20181231 & 0.0719 \\
2 & 20090101 & 20190630 & 0.0903 \\
3 & 20090101 & 20191231 & 0.0866 \\
4 & 20090101 & 20200630 & 0.1451 \\
\hline
\end{tabular}%
}
\caption{The RankIC on different rolling models}
\label{tab:exp}
\end{table}

\begin{figure*}[ht] 
    \centering
    \includegraphics[width=\textwidth]{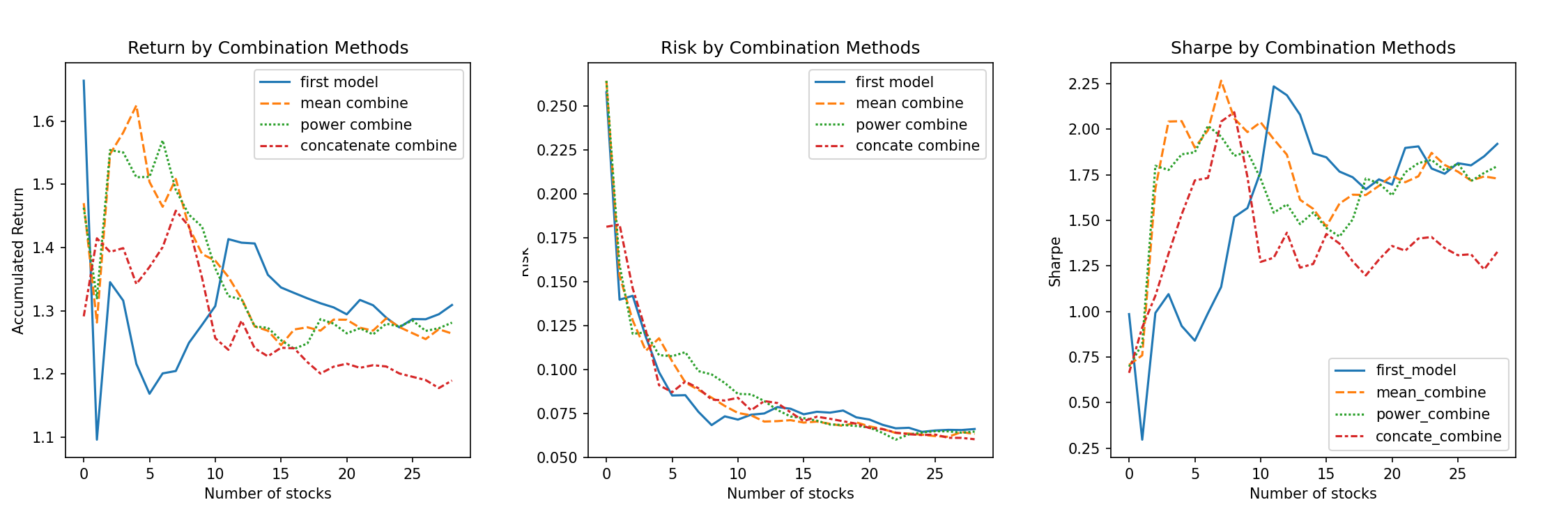}
    \caption{The return rate, risk and sharpe ratio by combination methods.}
    \label{fig:return}
\end{figure*}

In order to solve the timeliness problem and output reliable investment decisions from the models, we proposed the following 4 strategies to make investment decisions based on our model: single model strategy, mean combinatory strategy, concatenating combinatory strategy, and exponential combinatory strategy. The single model strategy simply uses the prediction model \#1 as the indicator. The mean and exponential combinatory strategy combines the results of all the models whose training set is earlier than the given testing time by averaging the predicted return rate or suming the exponent of the predicted return rate. Concatenating combinatory strategy uses the result from the model whose training set has the closest time to the testing.

As is demonstrated in figure~\ref{fig:return}, we computed the average return rate, risk(Standard Deviation of return rate), and sharpe ratio(annualized return rate/risk) of each strategy. The The mean combinatory strategy has the highest sharpe ratio and therefore is the most efficient and has the best balance between return and risk.

\begin{figure}[h] 
    \centering
    \includegraphics[width=\columnwidth]{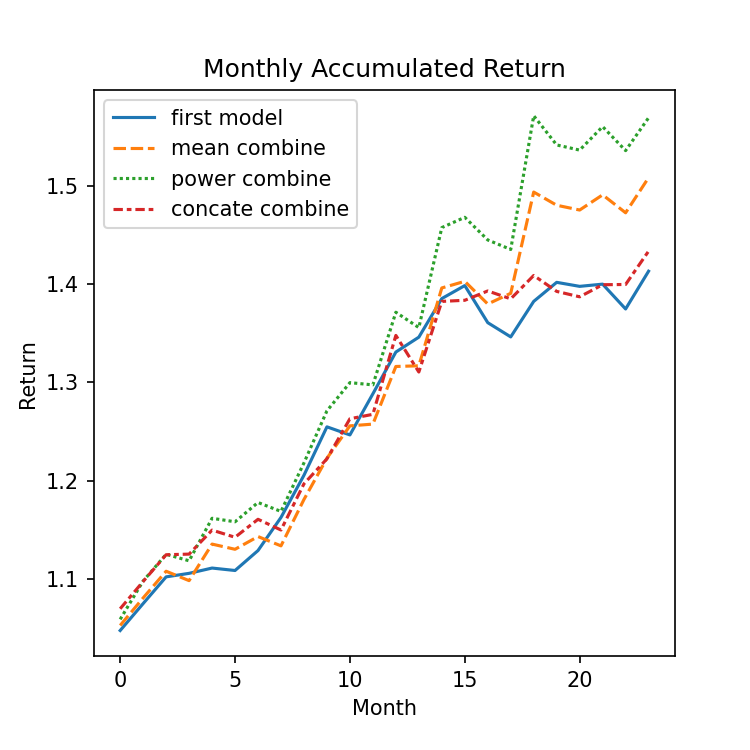}
    \caption{The monthly accumlated return on different strategies to make investment decisions based on ESAN.}
    \label{fig:monthly}
\end{figure}

As figure~\ref{fig:monthly} and table~\ref{tab:exp} shown, when selecting 8 stocks at each time, the mean combinatory strategy reaches 20\% annualized rate of return and 7.19\% RankIC. Such a RankIC score indicates significant correlation between the model prediction and actual return rate. Compared to the baseline, the strategy given by our models almost doubles the expected return rate and shows a higher correlation between prediction and true values.

\section{Conclusion}
In this paper, we propose a novel design, called ESAN.
NLP module and Time Series Model provide a general way of using nlp models in different finance strategies (and of course with different performances).
The main advantage of the NLP module is to guarantee the distinction and stability of temporal information. The time series model shows an extraordinary capacity to handle the temporal variation of NLP outputs. 
Our experiments on ESAN has proved to us its extraordinary learning ability due to its superb RankIC and rate of return. 

In practical applications, this whole system achieves impressive results on stock price prediction tasks. 
As future work, we are expecting to see more forms of such an nlp model on other stock price prediction tasks.

Our github repository link is as follows: https://github.com/Suntuo-493/DSCI544

Our video link is as follows:
https://www.youtube.com/watch?v=u0Ku0rl39IQ


\bibliographystyle{IEEEtran}

\clearpage
\begin{appendices}
    \section{Return Rate Results}
    
    \begin{table}[htbp]
\begin{tabular}{lllll}
\hline
\multicolumn{1}{c}{stock\_index} & \multicolumn{1}{c}{first\_model} & \multicolumn{1}{c}{mean\_combine} & \multicolumn{1}{c}{power\_combine} & concate\_combine\\ \hline
1          & 1.6642383202927367 & 1.4703567697190465 & 1.4631096327121487 & 1.2910990271791403 \\
2          & 1.0957938582756062 & 1.2813606988451771 & 1.3195929890835962 & 1.4147040362612073 \\
3          & 1.3452268237754619 & 1.548052759238498  & 1.5544022150220391 & 1.393121187287079  \\
4          & 1.3160788031430832 & 1.5821214955158003 & 1.5506800678647934 & 1.3991021323349306 \\
5          & 1.2158387384400242 & 1.6250574169230325 & 1.5114487952070264 & 1.3422317247414077 \\
6          & 1.1684928305066284 & 1.5039061223234902 & 1.5118291730457856 & 1.369239072695537  \\
7          & 1.2007697529140975 & 1.4646176618197568 & 1.569679625126854  & 1.4006906019030219 \\
8          & 1.2045408623498006 & 1.5086566960003767 & 1.4910968025529974 & 1.4583976554317448 \\
9          & 1.2491776665396392 & 1.432137476589687  & 1.452021631560068  & 1.433753135830949  \\
10         & 1.2778033946715126 & 1.3897032089385455 & 1.4330276225661294 & 1.3516720355042362 \\
11         & 1.3073804661421027 & 1.3792925459645362 & 1.367072165704628  & 1.2565926920719366 \\
12         & 1.4133287891690873 & 1.3534704624932643 & 1.3231289467966045 & 1.2381336566242296 \\
13         & 1.4077385982809902 & 1.3194205933387098 & 1.317791887910008  & 1.2843276760460707 \\
14         & 1.4064928561569219 & 1.2751055317736104 & 1.275064356340828  & 1.2404224050200392 \\
15         & 1.3569281216454798 & 1.2683175164351796 & 1.2729367597845107 & 1.227797633625205  \\
16         & 1.3368395983179668 & 1.2455298755437179 & 1.2533749392696727 & 1.2414008494239677 \\
17         & 1.3281249761911962 & 1.270276733827658  & 1.2396649762874425 & 1.2404530122821709 \\
18         & 1.3196694934011817 & 1.2737678863486408 & 1.2480201068714585 & 1.2187048085718222 \\
19         & 1.3118291696811277 & 1.2684721371376908 & 1.2866486817778855 & 1.200542316757104  \\
20         & 1.3052590873867467 & 1.285960494078427  & 1.2795114542083779 & 1.211424448479111  \\
21         & 1.294289905453318  & 1.2856995811274123 & 1.264243213752806  & 1.2160713552163867 \\
22         & 1.317120763916391  & 1.2730616799170593 & 1.2722139676161022 & 1.2095957897118752 \\
23         & 1.3086656540778836 & 1.2684972735875821 & 1.2625358443074244 & 1.2137189562600466 \\
24         & 1.2890235574328852 & 1.2879208718592698 & 1.2794329549860657 & 1.2117245203659337 \\
25         & 1.2735528704093237 & 1.2740640635563059 & 1.2749312763149114 & 1.200615091877732  \\
26         & 1.2868118841969405 & 1.264439516432548  & 1.2844522572229264 & 1.1953704713658375 \\
27         & 1.28650860140362   & 1.2550661600508448 & 1.2680610918104613 & 1.190187655741581  \\
28         & 1.294439395177561  & 1.2706644519184431 & 1.2722495580862008 & 1.1775042836295975 \\
29         & 1.3090992179436194 & 1.2638577857137316 & 1.2811319973795834 & 1.1894780367399276 \\ \hline
\end{tabular}
\end{table}

\clearpage
\section{Risk Results}
\begin{table}[htbp]
\begin{tabular}{lllll}
\hline
stock\_num & first\_model        & mean\_combine       & power\_combine      & concate\_combine    \\
\hline
1          & 0.25770639311159016 & 0.26427475561348407 & 0.26414145327788885 & 0.1812275784657754  \\
2          & 0.13976577827273323 & 0.15331957105241933 & 0.159539772920892   & 0.18254415758885872 \\
3          & 0.14197587595616096 & 0.12802271523985892 & 0.12035695016383433 & 0.14614426042180104 \\
4          & 0.11853200559242588 & 0.11075041607514144 & 0.12119433963811312 & 0.12220536464628161 \\
5          & 0.0985697583492108  & 0.1178166046785833  & 0.10821625464473615 & 0.09117284030698106 \\
6          & 0.08529784678929819 & 0.10470491625982936 & 0.10762912991516524 & 0.08719001267578563 \\
7          & 0.08547996246657635 & 0.09269205571123601 & 0.10992006172626573 & 0.0932388709347803  \\
8          & 0.07598785838142164 & 0.08849040648858833 & 0.09915796565189731 & 0.08941678082176212 \\
9          & 0.06845703100520194 & 0.08406321932532577 & 0.09725679605592402 & 0.08297155205339071 \\
10         & 0.07344140805796502 & 0.07936979433818113 & 0.0924021779547256  & 0.08243453893247023 \\
11         & 0.07159310586769717 & 0.0753726462413417  & 0.08621441569336778 & 0.08398783008754214 \\
12         & 0.07436245489499685 & 0.07402181207655245 & 0.08592907699264403 & 0.07689758984411234 \\
13         & 0.07507555842859298 & 0.07047788278894491 & 0.08217073070969699 & 0.08212894937294794 \\
14         & 0.07871608527081758 & 0.07071674078703559 & 0.0770746404649134  & 0.08100245714910573 \\
15         & 0.07777029045971515 & 0.07126543284829949 & 0.07336582560789998 & 0.07574175309593724 \\
16         & 0.07461145160925663 & 0.06988749065845329 & 0.07240701848271329 & 0.07077600730708071 \\
17         & 0.07604274145944505 & 0.0704629225563672  & 0.07099091010724973 & 0.07321970876566569 \\
18         & 0.07554515295994459 & 0.06919757919971084 & 0.06878587953059828 & 0.07205339479309661 \\
19         & 0.07675076037270877 & 0.06801553401712673 & 0.06852930945593148 & 0.07064303575100409 \\
20         & 0.07286540710618676 & 0.07000192851595377 & 0.06796843566910302 & 0.06926273158880936 \\
21         & 0.07162582219646617 & 0.06772287759544354 & 0.06708313589226576 & 0.06681622935318626 \\
22         & 0.06863084453414986 & 0.06626273736636484 & 0.06402441493006378 & 0.06616866020455077 \\
23         & 0.06663535459394712 & 0.06400721442357127 & 0.06017404379761824 & 0.0641930322025579  \\
24         & 0.0669351390993733  & 0.06363560018578103 & 0.06318109337258923 & 0.0632636300034126  \\
25         & 0.06461761922625592 & 0.06299981929821538 & 0.06424833492423497 & 0.0628038449961506  \\
26         & 0.0653938027364236  & 0.06219085128214357 & 0.06497874323767697 & 0.06306899221084598 \\
27         & 0.06575401883086832 & 0.061826031645346   & 0.06488840236216252 & 0.0612301340577514  \\
28         & 0.06564471325367667 & 0.06452571816351213 & 0.06415526684340973 & 0.06119232376106988 \\
29         & 0.0662456530545731  & 0.06341782900153548 & 0.06475860873047942 & 0.06041962577442166 \\ \hline
\end{tabular}
\end{table}

\clearpage
\section{Sharpe Ratio Results}
\begin{table}[htbp]
\begin{tabular}{lllll}
\hline
stock\_num & first\_model        & mean\_combine      & power\_combine     & concate\_combine   \\
\hline
1          & 0.9858471427993408  & 0.7071768427067514 & 0.6976770524515641 & 0.6635361181260699 \\
2          & 0.29689097520340657 & 0.7597707909532655 & 0.8221939173152187 & 0.9132520837449147 \\
3          & 0.9923272473957991  & 1.6744290649923057 & 1.7994484873881769 & 1.086352549622963  \\
4          & 1.0953395325826096  & 2.0421682873960227 & 1.7763248459317846 & 1.3172300864370654 \\
5          & 0.9206024076254481  & 2.044282345015508  & 1.8621821324366603 & 1.5328784138456353 \\
6          & 0.8400861302797162  & 1.8991446295823444 & 1.8735890315520554 & 1.7191542099679387 \\
7          & 0.9910449535029918  & 1.9939981114268983 & 2.018521068910085  & 1.7327346754933834 \\
8          & 1.1347500231761642  & 2.26613713146752   & 1.9595072559395186 & 2.041996937513278  \\
9          & 1.5182735923726998  & 2.0588915028662584 & 1.8537329444353452 & 2.0930672968331825 \\
10         & 1.5673593082075659  & 1.9843792903258932 & 1.8765904899058818 & 1.7385033167638435 \\
11         & 1.7670394163869072  & 2.038377100074234  & 1.729222775952829  & 1.272137976363214  \\
12         & 2.235060978021079   & 1.9452179676418395 & 1.5421877710732717 & 1.295073802973306  \\
13         & 2.186494040860585   & 1.8602566270223813 & 1.5879885604447934 & 1.4323326821695548 \\
14         & 2.0795532097722056  & 1.6129430728194851 & 1.4796836863765663 & 1.2405918279297168 \\
15         & 1.8681494956203106  & 1.5634831187649882 & 1.5432187664049999 & 1.2608475457321584 \\
16         & 1.8458240698221144  & 1.4666801988659293 & 1.4581946313073146 & 1.425295281450011  \\
17         & 1.7676553686890604  & 1.5921142193896964 & 1.4113395107305169 & 1.3726229815361393 \\
18         & 1.7367332271066245  & 1.6408450030535064 & 1.504403698619054  & 1.2752426043131433 \\
19         & 1.6704706620812995  & 1.6390743696451393 & 1.7296716136473376 & 1.1978982316809506 \\
20         & 1.725030878824845   & 1.6894858091912879 & 1.7033099134812226 & 1.2846985149758114 \\
21         & 1.6960462289534017  & 1.7448523122157682 & 1.6372830899372544 & 1.359494556083086  \\
22         & 1.8975471403757516  & 1.709378090085479  & 1.7639917569146633 & 1.3337255189851775 \\
23         & 1.905893317905654   & 1.741870953197067  & 1.814198888181504  & 1.4004316673462185 \\
24         & 1.784568163668115   & 1.8704091019095663 & 1.8318911521611752 & 1.4084175342232075 \\
25         & 1.7558516939817301  & 1.8040936008348545 & 1.77427803032462   & 1.3478854451306046 \\
26         & 1.8135702625695609  & 1.7673125272103432 & 1.8111178247595257 & 1.3087892925576095 \\
27         & 1.8018530047647352  & 1.7184719295102904 & 1.715598213148161  & 1.3139844839632706 \\
28         & 1.851455613275641   & 1.7409474020468159 & 1.7606095847514558 & 1.2309204243724474 \\
29         & 1.9196098388072131  & 1.729540314646005  & 1.7974314652355754 & 1.326871678039145 \\ \hline
\end{tabular}
\end{table}

\clearpage
\section{Monthly Accumulated Return Results}

\begin{table}[htbp]
\begin{tabular}{lllll}
\hline
month & first\_model       & mean\_combine      & power\_combine     & concate\_combine   \\
\hline
1     & 1.0472151759674282 & 1.0523720302873754 & 1.058598031063293  & 1.0693931256413054 \\
2     & 1.0746726102892314 & 1.0804142711536389 & 1.0976754950074612 & 1.097207035798338  \\
3     & 1.1018748686998003 & 1.1076165295642078 & 1.1248777534180303 & 1.1244092942089068 \\
4     & 1.1054703519991353 & 1.0980367740550159 & 1.1182549654216989 & 1.1250611555795587 \\
5     & 1.1108587274725985 & 1.1352705228790794 & 1.1615579955381756 & 1.149525395264624  \\
6     & 1.1083696088833217 & 1.1299735691577268 & 1.1580104431572438 & 1.1422195646949358 \\
7     & 1.1287034608632536 & 1.1429513918498615 & 1.1776615232437677 & 1.160539475154716  \\
8     & 1.162443395689312  & 1.133464753953483  & 1.168555091920247  & 1.149613306918031  \\
9     & 1.2054301482579652 & 1.180748597836285  & 1.2180127481812186 & 1.196707427698325  \\
10    & 1.254745033763182  & 1.2227914024154116 & 1.271056023373988  & 1.222171869884597  \\
11    & 1.2465037155605323 & 1.2557482404956273 & 1.2998045643693013 & 1.2629915129297185 \\
12    & 1.2884102970771152 & 1.2575409742217536 & 1.2974953046557207 & 1.2674965839963683 \\
13    & 1.3309569518348563 & 1.3162558699716564 & 1.3715163057073607 & 1.3479094381893022 \\
14    & 1.3461275018824952 & 1.316952669885945  & 1.3558718413383226 & 1.310599548113867  \\
15    & 1.3854592780939439 & 1.3962639161278418 & 1.4577518035756027 & 1.3825695069682882 \\
16    & 1.3986879866167488 & 1.4030056373346111 & 1.468259501685252  & 1.3837076175313634 \\
17    & 1.360895640043496  & 1.37990855698046   & 1.445184871624725  & 1.3932234457901371 \\
18    & 1.346394204985476  & 1.390816904793632  & 1.4354988925645809 & 1.385144532613459  \\
19    & 1.382577311923002  & 1.4939824264471706 & 1.5717510850363117 & 1.408984799484386  \\
20    & 1.4019843957220735 & 1.4805415835275788 & 1.5419917872888598 & 1.3927633377253887 \\
21    & 1.3978960153079139 & 1.4756704996957302 & 1.5367282532296234 & 1.387241586259008  \\
22    & 1.4001518453448702 & 1.491151004025656  & 1.5606606330216557 & 1.3995653002281059 \\
23    & 1.3748486555583856 & 1.4728630019906355 & 1.5362062042887386 & 1.3998204404445536 \\
24    & 1.4133287891690873 & 1.5086566960003767 & 1.569679625126854  & 1.433753135830949 \\ \hline
\end{tabular}
\end{table}
\end{appendices}

\end{document}